\pdfoutput=1

\documentclass[11pt]{article}

\usepackage[preprint]{acl}

\usepackage{times}
\usepackage{latexsym}

\usepackage[T1]{fontenc}

\usepackage[utf8]{inputenc}

\usepackage{microtype}

\usepackage{inconsolata}

\usepackage{graphicx}

\title{Patronizing and Condescending Language Detection in English and Mandarin}

\author{Xinyi Vicky Xiang \\
  University of Washington \\
  \texttt{xinyix7@uw.edu} \\\And
  Catherine Qingyun Wang \\
  University of Washington \\
  \texttt{qywang@uw.edu} \\\And
  Paula Cortés Lemos \\
  University of Washington \\
  \texttt{mpcortes@uw.edu} \\}

\begin{document}
\maketitle
\begin{abstract}

Patronizing and condescending language (PCL) is a form of harmful language that belittles and assumes pity for vulnerable groups, reinforcing negative societal stereotypes. However, unlike overt hate speech, it is often difficult to detect, even for human judges. This paper proposes a system for the automated detection of PCL. Our first approach uses static word embeddings and a convolutional neural network for classification, while our second approach uses pre-trained BERT embeddings fine-tuned for the PCL task. Our second approach was considerably more successful, and achieved an F1 score of 0.528 over the development set. 

\end{abstract}

\section{Introduction}

Patronizing and condescending language (PCL) is often expressed unconsciously, but is nonetheless harmful towards vulnerable communities. Speaking about or generalizing entire communities in a pitying way can lead to the reinforcement of harmful stereotypes and normalize discrimination. The \textit{Don't Patronize Me!} dataset \cite{perez-almendros-etal-2020-dont} is a large-scale dataset for PCL detection and classification in English. However, PCL is usually subtler and often unconscious compared to other forms of hate speech or sarcasm, making it difficult to detect even for human judges. Nevertheless, previous work in this area (specifically in the SemEval 2022 task) has been able to achieve encouraging results that demonstrate a PCL-detection model is viable, even if currently it has achieved slightly limited results.\\

Moreover, creating a well-performing PCL-detection model could have a positive impact in texts that deal with sensitive, potentially offensive topics, as it could guide authors to rephrase ideas conveyed (often unintentionally) in a patronizing way.\\

With this in mind, we propose and implement a system for detecting PCL, first using static word embeddings and convolutional neural networks (CNNs), and using a transformers-based approach by fine-tuning pre-trained BERT embeddings. 

\section{Task Description}

Our primary task comes from task 4 of SemEval 2022, which focuses on the detection of Patronizing and Condescending Language (PCL). The task describes PCL as language that `reveals a superior attitude towards others' and is `often unconscious but harmful and discriminative' \cite{perez-almendros-etal-2022-semeval}. Unlike toxic or offensive language, PCL is not overtly aggressive or inflammatory, but nonetheless furthers power imbalances and harmful stereotypes towards vulnerable communities. For example, the following positive example from the \textit{Don't Patronize Me!} dataset shows how patronizing and condescending language presents superficial solutions and establishes privileged groups as `saviors' of those in need:

\begin{quote}
“Norberto Quisumbing Jr. of the Norkis Group of Companies has a challenge for families who can spare some of what they have: why not adopt poor families and help them break the cycle of poverty?”
\end{quote}

The data set used for this task is the \textit{Don't Patronize Me!} data set presented in \citet{perez-almendros-etal-2020-dont}, an annotated corpus of PCL towards vulnerable communities. The authors define seven subcategories of PCL: 

\begin{itemize}
    \item Unbalanced power relations: the author positions themselves as a savior of those in need
    \item Shallow solution: the author presents an overly simple solution as though the hardship were superficial
    \item Presupposition: the author makes assumptions or stereotypes about a vulnerable community
    \item Authority voice: the author presents advice as an authority about how to overcome underprivileged situations
    \item Metaphor: the author describes difficult situations in softer, poetic ways
    \item Compassion: the author uses flowery wording to describe the group with pity
    \item The poorer, the merrier: the author praises the hardships as beautiful
\end{itemize}

The SemEval 2022 task has two subtasks: first, the idenfication of PCL, a binary classification task; and second, the identification of which of the seven categories a given instance of PCL belongs to, a multi-class classification task. 

Our primary task will focus on subtask 1. This task involves analyzing English text to detect the affect of condescension, a binary classification task. The task will involve text data only, and the genre of said text will be excerpts from news articles. 

Our adaptation task will adapt this task for another language, Mandarin Chinese. We will use the CCPC (Chinese Corpus for Patronizing and Condescending Language) corpus, a corpus of discourse from Weibo and Zhihu platforms \cite{10.1007/978-3-031-44696-2_50}. This corpus additionally annotates entries with `PCL toxicity strength', but can still be used for the binary classification task, distinguishing between entries that are labeled as not toxic and entries that are labeled with any level of toxicity.



\section{System Overview}

Our initial system features two separate approaches to the PCL detection task. Our first approach tackles the classification task using static embeddings to represent the text and neural networks to perform classifications. We use pre-trained GloVe embeddings as input and, following other attempts at the task using CNNs (such as \citet{siino-etal-2022-mcrock}) and experimentation, incorporate dropout and pooling in our initial network architecture. Although the approach outperforms a random baseline, the overall accuracy over our devset is quite poor. Thus, we consider a second approach, a transformer-based approach in which we fine-tune a pretrained BERT for classification model over our data set. This approach saw considerably more success, with an F1 score of 0.53 over our devset. 

Our end-to-end system includes data preprocessing by removing URLs, html tags, punctuation, emojis, and stopwords. We also balance the data, as the initial dataset is heavily skewed towards the negative class. Finally, our system evaluates outputs compared to gold standard labels over a development set, producing precision, recall, and F1 scores.

\section{Approach}

In this section, we review our approach and the components of our CNN and transformer models. We will first briefly overview our data preprocessing, then go into detail on the architecture of each model.


\subsection{Dataset Composition}

For our primary task, the data from the \textit{Don't Patronize Me!} dataset contains 10,469 entries. To avoid tuning to the test data during our development phase, we partition this data into training and development sets, resvering 10\% of data entries for the devset. 


The data set labels each entry with an integer value between 0 and 4, inclusive. These represent the `level' of PCL determined by the human annotators, with 0 being the least and 4 being the most condescending. In accordance with the methodology outlined in \cite{perez-almendros-etal-2022-semeval}, all labels were initially transformed into binary states. Labels originally designated as 0 and 1 were categorized as non-PCL cases and subsequently relabeled as 0. Conversely, labels originally designated as 2, 3, or 4 were identified as PCL cases and subsequently relabeled as 1. 

\subsection{Text Preprocessing}

During our text preprocessing, we utilized NLTK tokenizer and BERT's default tokenizer to perform tokenization. Subsequently, we replaced any characters that are not English letters or punctuation with a blank space character. Furthermore, we eliminated any URLs and HTML tags to enhance clarity. Lastly, we removed punctuation characters from a predefined string encompassing common English punctuation symbols. The removal of these features in sentiment analysis has been proven to generally have a positive impact on results, as it allows the data to be less noisy, so the system can focus on high-impact features \cite{alamyao}.

Additionally, for our BERT model, we also pad our sentences with \verb|[CLS]| and \verb|[SEP]| tokens to each sentence for encoding purposes. Specifically, \verb|[CLS]| token is prepended to the start and  \verb|[SEP]| token is appended to the end. This then creates attention masks to differentiate between padding and non-padding tokens.



\subsection{Data balancing}

\subsubsection{Initial data balancing}

The training set for our primary task is an unbalanced data set, with 90.5\% of data entries being negative samples. This posed problems for our convolutional model, as it is able to achieve high validation accuracy by merely predicting 0 (not PCL) for every entry, resulting in the model weights quickly zeroing out over training epochs. To combat this, our initial model balances the data by undersampling the negative samples to create a dataset that has equal positive and negative samples. However, this dataset is much smaller, with just over 1,000 samples rather than over 10,000, and as a result, we lose a lot of the information in those negative samples. For this reason, we experimented with two data augmentation methods, explained in the following section.




\subsection{Revised data balancing}

In our initial system, we used a naive method of random undersampling to deal with class imbalance in our data. However, this method revealed to be insufficient, especially because the resulting undersampled dataset was so small that we did not feel it would be sufficient for fine-tuning our BERT model. By undersampling, we lose a lot of the information that would be important for training.

Therefore, in our revised model, we take a data augmentation approach. Instead of undersampling the negative samples, we use two different data augmentation approaches to increase the amount of positive samples, thus balancing the two classes. We experimented with both Random Deletion and Synonym Replacement as data augmentation methods. However, the results showed that with our BERT model, they did not improve performance, so we may not necessarily implement them in our final model. 

\subsubsection{Random Deletion}

We applied random deletion over the set of training samples to create more artificial sentences for the positive class (sentences with PCL). As explained by \cite{CHAO2023111038}, random deletion is capable of preserving the semantic characteristics of data, as keywords normally represent just a small fraction of the sentence and are unlikely to be deleted; additionally, the data it creates is more diverse and can thus make the model more generalizable.

We set our script to create new sentences by 1) iterating through the positive class samples, 2) selecting a set of \(n\) random words for each sentence, without replacement, where \(n\)  is one of our model's hyperparameters, 3) creating \(n\) new sentences per sample, where each of them is identical to the original sample except for one deleted word. For example, with \(n=2\):

\begin{quote}
    \textbf{Sample sentence:} "depicts demonstrations refugees border post , catastrophic living conditions desperate attempt several hundred cross river kilometers camp get macedonia march"
    
    \textbf{Artificial sentence \#1:}
     "depicts demonstrations refugees border post , catastrophic living conditions desperate attempt several hundred cross river kilometres camp get macedonia" [\textit{march} was deleted]
     
     \textbf{Artificial sentence \#2:}
     "depicts demonstrations refugees border , catastrophic living conditions desperate attempt several hundred cross river kilometres camp get macedonia march" [\textit{post} was deleted]
     
\end{quote}

Initially, we set the model to create 9 additional sentences per positive sample, as the class imbalance had an approximate ratio of 1:10 positive to negative sentences. After this process, we have a ratio of approximately 1:1 positive to negative samples. However, as this did not seem to improve the performance, we experimented with reducing this hyperparameter to 3 new sentences per original sample (resulting in 2:5 positive to negative samples), which yielded better results.





\subsubsection{Synonym Replacement}
Similarly to random deletion, we implement synonym replacement to create more artificial sentences to augment our positive class samples. We base our synonym substitution on the method used by \cite{vazquez-ramos-etal-2022-i2c}. Specifically, we use WordNet (imported from NLTK) to access a list of synonyms for each word.

In an analogous manner to the approach in our random deletion method, we perform synonym substitution by 1) selecting a random word index in each sentence to replace it with a synonym, 2) selecting a random synonym from the list of synonyms for that word, 3) replacing it in the original sample, thus creating a new sentence. We repeat this process \(n\) times per positive sample, where \(n\) is another hyperparameter of our model. For example, one instance of synonym replacement could look like the following:

\begin{quote}

    \textbf{Sample sentence:} "least  hungry \textbf{traumatised} refugees sought refuge bangladesh since october"
    
    \textbf{Artificial sentence:}
     "least  hungry \textbf{shock} refugees sought refuge bangladesh since october" [\textit{traumatised} was replaced by its synonym \textit{shock}]
     
\end{quote}

We set our model to create 3 new sentences per positive sample in the training set. This approach produced better results in comparison to the random deletion method.

\subsection{System Architecture}

\subsubsection{CNN with static embeddings}

For our initial model, we use static GloVe embeddings to represent the text data. Specifically, we use the average vector sum of the tokens in each data entry to create an input vector for each data entry, following the Sentence2Vec method. This allows the representation for each entry to be comprised of its token embeddings while being normalized for sentence length. 

We use PyTorch dataloaders to create tensors for our training and validation set, which are randomly split 80-20. After experimenting with different model architectures, our most recent model uses cross-entropy loss and the Adam optimizer. Following the findings from \cite{siino-etal-2022-mcrock}, which demonstrated the success of convolutional layers involving different kernel sizes as well as dropout and pooling, we experimented with the kernel size for the \textit{Conv1D} layer filters and applied dropout and max pooling. With some experimentation, we also found 0.0001 to be the best learning rate for our model. The following graphs show the loss and training and validation accuracy of this model over 50 epochs. 

\subsubsection{BERT model}

Condescending language is often subtle, context-dependent, and can be challenging to detect accurately. The low baseline scores for this task and the poor performance of our initial model suggest that static embeddings may not be enough to capture the dependent relationships between words and the phrasal subtleties of condescending language. For these reasons, we found BERT (Bidirectional Encoder Representations from Transformers) suitable for the PCL detection task as they offer advantages in terms of contextual understandings. In addition, BERT's ability to represent the semantics of language helps in identifying condescending expressions even when they are not overtly stated. 

Given the similarity of PCL detection to other subtle affect detection tasks such as hate speech detection, and the popularity of transformer models such as BERT for these tasks, we chose to adapt a transfer learning approach by fine-tuning a pre-trained BERT model on the PCL detection dataset. In particular, this model is trained with self-supervision over unlabeled text with the masked language model and next sentence prediction tasks \cite{devlin-etal-2019-bert}. 

Specifically, we employed BERT-base-uncased, which was pre-trained in English with 12 layers, 768 as its hidden size, 12 heads, and 110M parameters. BERT utilizes almost the same architectures for pre-training and fine-tuning for different tasks; the only difference is the output layer, which can vary depending on the goal of the model. We fine-tuned our system for binary classification with a batch size of 16 and 4 epochs, within the range recommended by \citet{devlin-etal-2019-bert}. However, after training, we noticed that for the last two epochs, loss was continuing to drop with little change in validation accuracy. Thus, to avoid possible overfitting, we decided to retrain the model with only 2 epochs, and found that we achieved better results over our devset. 

\subsubsection{Revised BERT model}

In our initial system, we lowered the number of training epochs from 4 to 2 to prevent overfitting. In our revised system, we also experiment with other hyperparameters, such as learning rate and the number of samples generated in data augmentation, as well as with different preprocessing methods. For example, we experimented with linear learning rate scheduling with different initial learning rate values, and with the epsilon value for the AdamW optimizer. We also experimented with removing vs. not removing punctuation, emoji, and stopwords in our preprocessing phase. 

After experimenting, we found that the system performed better when the initial learning rate was raised to $5 \times 10^{-5}$. We thus use this initial learning rate with the linear learning rate scheduler, and an AdamW optimizer with an epsilon value of \(1 \times 10^{-8}\). We found that for the BERT model specifically, removing stopwords resulted in better performance, and removing teminal punctuations and tone punctuations (emojis) did not significantly impact performance. In fact, results were actually marginally better without removing punctuation, so we decide to leave in both types of aforementioned punctuations. Finally, we found that only generating 3 samples per real sample in our data augmentation methods seemed to be the best. Although the classes do not get perfectly balanced, we saw that they provided empirically better results. However, because it seemed that the data augmentation methods did not improve upon the initial results, we are currently not implementing them in our final system.

\section{Results}

\begin{table*}[t]
\centering
\begin{tabular}{llll}
\hline
\textbf{Model} & \textbf{Precision} & \textbf{Recall} & \textbf{F1 Score}\\
\hline
Linear NN & 0.129 & 0.038 & 0.058 \\
Conv1D NN (with undersampling) & 0.084 & 0.343 & 0.135\\
BERT (4 epochs) & 0.589 & 0.306 & 0.403 \\
BERT (2 epochs) & 0.565 & 0.495 & 0.528 \\
\hline
\end{tabular}
\caption{\label{tab:initial-results}
Precision, recall, and F1 score for each model in our initial system over the development set. 
}
\end{table*}

\begin{table*}[t]
\centering
\begin{tabular}{llll}
\hline
\textbf{Data Augmentation Method} & \textbf{Precision} & \textbf{Recall} & \textbf{F1 Score}\\
\hline
Random Deletion (9 samples) & 0.103 & 1.0 & 0.187 \\
Random Deletion (3 samples) & 0.429 & 0.438 & 0.433\\
Synonym Replacement (3 samples) & 0.475 & 0.372 & 0.417 \\
\hline
\end{tabular}
\caption{\label{tab:augmentation-results}
Precision, recall, and F1 score for each model type in our revised system over the development set. 
}
\end{table*}

This section will present the initial results of our system over the aforementioned development set that we created. The development set has 1,046 entries, with 105 positive samples, thus having a similar proportion of negative to positive samples as in the complete training set. 

Our system is set up to output predictions in the form of a text file with one prediction, 0 or 1, on each line. We use our own script, which is adapted from the official scorer of the SemEval-2022 task, to perform our evaluation (we are unable to use the official scorer for now, as it compares the predictions to the test set, while we are currently evaluating over our development set).

The official scorer produces three scores for each subtask: precision, recall, and F1 score, comparing the predicted labels to the ground truth labels, with 1 being the positive class. Table 1 shows the performance of our system with each of the following model specifications. (The results of our CNN model without undersampling are not included; because the model predicted the negative class for every single entry, the precision and recall are both 0.)

Ultimately, our static embedding based models performed quite poorly, with both the linear and convolutional NN used for classification. The fine-tuned BERT model performed the best, with an improvement in F1 score after training for only 2 epochs. We have not tested the BERT model with undersampling due to concerns about the undersampled dataset being too small to fine-tune the BERT model; in future iterations, we plan on experimenting with other methods of data augmentation for the BERT model.

Table 2 shows the results of our experimentation with data augmentation methods and hyperparameter tuning. We experimented with the number of synthetic samples generated per real sample of the minority class, and found that although there was a 9-to-1 class imbalance, generating 9 samples per real sample resulted in the model overly predicting the positive class, as shown in the first row of the table. 

Specifically, the results shown in the table are with 2 training epochs and an initial learning rate of $5 \times 10^{-5}$, which we found to yield the empirically best results. Unlike the initial system, we also chose not to remove terminal punctuations and tone punctuations (emojis) during preprocessing, as that also resulted in marginally better results.  

As we can see, neither data augmentation method outperformed the initial results. Looking ahead, we consider two enhancement pathways structure wise for the prediction system and data wise: (1) system wise, we aim to leverage ensemble methods which combines system of BERT, Roberta, Distilbert, Roberta large, Albert; then potentially, map from existing language sources to other languages and back. (2) data wise, one approach we have in mind is to enhance strategies for class variations among data;
additionally, we hope to consider transformation techniques that have not been adopted for this deliverable, so far, we have implemented Synonym Replacement and Random Deletion but have not used Random Insertion and Random Swap. It remains to be discovered how the currently unutilized augmentation methods would affect accuracy and average training losses.

\section{Discussion}

\begin{table*}[t]
\centering
\begin{tabular}{p{9cm} p{2cm} p{2cm}}
\hline
\textbf{Text} & \textbf{Ground Truth Label} & \textbf{Predicted Label}\\
\hline
\small "They follow a crippling El Nino-triggered drought which scorched much of the region last year , hitting crop production and leaving millions in need of food aid ." & 0 & 1 \\
\small "If they couldn't , automated "" robo-debt "" letters told them to pay up , in an inversion of the usual onus of proof . It worked like extortion . Some of the victims were vulnerable , some couldn't cope ." & 0 & 1 \\
\small According to Betty-Ann Blaine , executive director of Hear the Children 's Cry , deterioration of family life ; instability/shifting households ; crippling levels of poverty ; lack of adequate social support systems and heartbreaking levels of hopelessness are key factors leading to children running away from homes .& 1 & 0\\
\small "Homeless children are "" living like refugees in their own country "" , deprived of the ability to make choices about when they eat , and in many cases suffering high levels of anxiety about their parents ' health ." & 1 & 0  \\

\hline
\end{tabular}
\caption{\label{tab:bert-errors}
Some examples of errors made by our fine-tuned BERT system.}
\end{table*}

\subsection{Initial system}

Compared to the random baseline and the best-performing systems at SemEval-2022, our first implementation performs extremely poorly, while our second implementation achieved substantial improvement and performs reasonably well on the binary classification task. Nonetheless, there is still much room for improvement; this section will discuss some of the errors our system made and how the system can be improved in future iterations.

The first issue is the unbalanced data issue. Before we implemented undersampling, our CNN system predicted `not PCL' for every entry on the dev test, suggesting that the unbalanced data was allowing the system to achieve very high validation accuracy without even considering the positive class. Of course, this defeats the purpose of the system, as the goal of PCL detection is specifically to detect positive cases when they occur. However, our random undersampling approach was not particularly successful, either. This approach resulted in a much smaller dataset than we started with, meaning that a lot of helpful data is simply lost. Because convolutional neural networks often require large amounts of data for optimal performance, we do not want to be throwing away large amounts of training data that could be useful for the inference task. 

Another issue is the issue of overfitting during training. Though our CNN system was able to achieve high training and validation accuracy and low loss over 50 epochs, the F1 score over the development set was still poor. We suspect that this also somewhat has to do with the unbalanced data; even when training on the balanced undersampled data, the system will still perform poorly over the dev set because it is also unbalanced. Moreover, the small training set size as a result of the undersampling also makes it prone to overfitting. The overfitting issue could also be alleviated by cross-validating across multiple validation and development sets, rather than only using a single split.

In future iterations, we believe we can improve upon the CNN system by improving the data augmentation method and by conducting further experiments with the model architecture, including the inclusion of dense layers, or a multi-channel CNN model as seen in \citet{siino-etal-2022-mcrock}. Moreover, alternate vector representations for each input data point could be considered; the current method of using static word vectors and aggregating them to create a vector representation of each data point could be improved using document vector representations or other feature sets. 

However, we focus our attention on our BERT model, which is currently achieving much better performance. We have already run tests over different epoch counts to avoid overfitting, but further experimentation will be needed to tune other hyperparameters such as the learning rate scheduler. 

Our preliminary results show that our model achieves a better precision than recall score over our development set. Table 3 shows some examples of errors made by our fine-tuned BERT system. As we can see, the difficulty in PCL detection lies partially in that many of the passages that contain PCL and those that do not use many of the same phrases. For example, in these examples, phrases like `crippling', `vulnerable', or `anxiety' can be used either factually to denote a specific state, or in a flowery, pitying way. The challenge is in differentiating between sentences about the same concepts based on the tone with which the author describes these concepts, and that distinction is not always made by the inclusion or exclusion of specific phrases. In future iterations, we may consider ensemble methods that may help us mitigate some of the brittleness of the current model.



\subsection{Revised system}

As previously shown, the implementation of data augmentation to help with class balancing did not improve upon the original model. This may have to do with the dev (and test) set itself being imbalanced as well, resulting in the model learning different base probabilities. Additionally, it is possible that due to the complexity of PCL, word deletion and synonym replacement cause sentences to lose some of the nuance that made them condescending or patronizing. The subtle distinctions among synonyms for a concept might lead to the attenuation of any patronizing undertones within a sentence, even if its core semantic meaning remains unchanged after the substitution. With these considerations, we believe it could be useful to implement with more subtle data augmentation techniques (creating less artificial sentences), or directly foregoing data augmentation altogether.

Moreover, the model as it stands is able to recognize phrases that are hallmarks of PCL, but still falls short of recognizing the context needed to properly assign PCL. Part of the issue has to do with the subjective nature of PCL and the fact that the training data frequently includes both sentences that show the opinion of the authors \textit{and} sentences that merely include a quote, such as in the below examples:

\begin{quote}
    Norberto Quisumbing Jr . of the Norkis Group of Companies has a challenge for families who can spare some of what they have : why not adopt poor families and help them break the cycle of poverty ?
\end{quote}

\begin{quote}
    Bombarded by schizophrenia , addiction and homelessness , you might say that Eoghan O'Driscoll has been to hell and back . But he is finding a new balance through painting .
\end{quote}

The model may struggle to pick up on these kinds of features. Moreover, the dataset itself is often inconsistent on distinguishing between cases of PCL in the news article itself vs. within quotes, making it even more difficult.

Future improvements may require task-specific adjustments to the pre-trained BERT model (such as adding certain additional layers) or ensemble methods which construct a combined system of language representation models. 

\subsection{System Resource Utilization}

We are currently developing and training our system locally, using CUDA on a local GPU. The initial training time for our BERT model was over 4 hours, but subsequent training (for each of our iterative changes) has required approximately 7:50 minutes per epoch, resulting in approximately 15 minutes to train two epochs, with an additional 2-3 minutes for validation. Below is a sample output from our training script, which times the training and validation process:

\begin{verbatim}
======== Epoch 1 / 2 ========
Training...

  Average training loss: 0.65
  Training epoch took: 0:07:54

Running Validation...
  Accuracy: 0.91

======== Epoch 2 / 2 ========
Training...

  Average training loss: 0.64
  Training epoch took: 0:07:52

Running Validation...
  Accuracy: 0.91

Training complete!
Total training took 0:16:30 (h:mm:ss)
\end{verbatim}

\section{Ethical Considerations}

Though the intent of PCL detection is to identify harmful language and mitigate its usage, there are ethical considerations for the usage of automatic PCL detection. Because the question of what is and is not PCL can be difficult for human evaluators and is sometimes subjective, an automated PCL detection system could be abused against authors who themselves are from underprivileged communities trying to express their experiences or show solidarity. The designation of certain texts as PCL without context of the writer's background could be harmful to well-intentioned authors who are just trying to share their experiences.\\

In general, given that PCL can be characterized as a form of hate speech, implementing the system without sufficient awareness of its limitations could lead to potential serious harmful accusations. For this reason, we suggest careful approaches when using this system, as well as ensuring sufficient monitoring to mitigate any risks of misuse or misinterpretation.

\bibliography{custom}

\end{document}